# Intelligence as Managed Autonomy: Failure, Escalation, and Governance for Agentic AI Systems

Srini Ramaswamy, Ph.D.
DNRS.ai USA
srini@computer.org

***Abstract -*** As autonomous and agentic AI systems scale in robotic and human-machine environments, managing hallucination and persistent but unjustified action remains an open challenge. Rather than attributing these failures solely to model or alignment limitations, this paper explores the architectural vulnerability of unbounded autonomy - the presumption that an agent should continue operating regardless of rising uncertainty. It introduces a theory of managed autonomy that defines intelligent behavior through the formal capacity to detect epistemic drift, suspend reasoning, attempt recovery, and ultimately surrender control when reliability diminishes. We instantiate this theory via the SMARt (Self-Managing Multi-tier Autonomous Reasoning with Regulated/Revoked transitions) model, a four-layer framework featuring Stable, Meta-cognitive, Assisted, and Regulated states. By developing a timed, guarded Petri net formulation, we establish theoretically bounded properties for the system, demonstrating how architecture can formally mandate escalation, constrain invalid outputs, and ensure governance reachability under specified conditions. We further analyze how incorporating domain-specific trigger sets across varied operational settings (e.g., healthcare, robotics, etc.) can systematically preserve safety, assuming completeness and soundness criteria are met. Because these triggers are designed to be adaptive, the SMARt model accommodates the safe, controlled expansion of an agent's operational scope over time. We conclude that formalizing failure management within the autonomy lifecycle is a crucial step toward realizing reliable and governed artificial intelligence.

***Keywords:*** *Failure-Managed Autonomy, State-Based Agentic Intelligence, Hallucination and Uncertainty Suppression, Autonomous Escalation and Governance, Hierarchical Recoverability Architecture, Intelligent Control and Petri Net Formalism, Safe Autonomous and Robotic Systems, Domain-Grounded General Intelligence.*

## I. INTRODUCTION

Autonomous and agentic AI systems are increasingly deployed in complex environments where operational failure is not exceptional but inevitable. These systems plan, act, reflect, and coordinate across tools, models, and agents, often with minimal human oversight. Despite advances in reasoning capability, a persistent challenge remains: these architectures frequently continue to execute and output confidently under conditions where they lack sufficient grounding. This failure mode, commonly observed as hallucination[1] or overconfidence in large language model (LLM)-based agents, is typically attributed to deficiencies in training data, model alignment, or prompt design. However, we propose that ungrounded continuity can be productively analyzed as a failure of autonomy control rather than strictly a language generation error. Specifically, such outputs arise when autonomous reasoning is permitted to persist beyond its verifiable epistemic validity.

### *A. The Limits of Unbounded Autonomy*

Most contemporary agentic AI architectures operate under the implicit assumption that autonomy, once granted, should persist indefinitely unless explicitly interrupted by external forces. Mechanisms such as reflection, self-critique, and multi-agent debate are largely implemented as procedural heuristics rather than enforceable constraints. Because failure to resolve uncertainty within these loops rarely forces a system to explicitly suspend operation, agents remain structurally biased toward continuous execution. Consequently, behaviors resembling hallucination emerge not simply as anomalies, but as predictable artifacts of an architecture lacking formalized exits from autonomous operation. While internal reasoning models have grown highly sophisticated, we argue that mitigating ungrounded outputs requires explicit, enforceable boundaries on the agent's structural authority to act.

### *B. Intelligence as Failure Management and the SMARt Model*

To address this structural gap, we introduce the SMARt (Self-Managing Multi-tier Autonomous Reasoning with Regulated transitions) Autonomy Model. Rather than defining intelligence solely by the capacity for continuous task execution, SMARt formalizes the ability to manage failure: to detect epistemic degradation, suspend autonomous action, initiate recovery, and relinquish control when necessary. SMARt operates as a state-based framework comprising four mutually exclusive modes: *(i)* Stable Autonomous Reasoning (S): The system operates within verified epistemic bounds; this is the only state where authoritative external output is permitted. *(ii)* Meta-cognitive Local Recovery (M): Autonomous execution is suspended for self-diagnosis and internal repair. *(iii)* Assisted Mutual Recovery (A): The system engages external agents or tools to resolve knowledge gaps it cannot self-correct. *(iv)* Regulated Transition to External Control (Rt): Autonomy is explicitly surrendered to governance or human oversight. SMARt grounds state transitions in domain-specific, measurable evidence signals - such as retrieval-LLM divergence, multi-agent disagreement, or chain-of-thought entropy - rather than abstract truth predicates. By utilizing observable metrics as strict state-transition triggers, the framework transforms the management of uncertainty into a regulable process governed by explicit stopping rules.

---

[1] For the purposes of this paper, we define ungrounded generation of recommendations (ex. text) as hallucinations.

### C. Formal Grounding and Scope of Contributions

Building on classical hierarchical intelligent control, we formalize the SMARt model using timed, guarded, hierarchical Petri nets. This mathematical realization provides a formal substrate intended to make corrigibility and governance enforceable architectural constraints. This paper makes the following contributions: (i) a theoretical reframing of hallucination as a failure of autonomy management; (ii) the specification of the SMARt Autonomy Model; (iii) a formal Petri net realization of the framework; and (iv) theoretical propositions demonstrating structurally bounded autonomous reasoning and mandatory escalation under uncertainty. To properly situate these contributions, it is essential to clarify the scope of this work. The SMARt model is introduced primarily as a conceptual and theoretically grounded autonomy framework rather than an empirically validated systems architecture. While subsequent sections outline conceptual pathways for real-world integration, the primary objective of this paper is to establish the mathematical properties, state transitions, and structural boundaries required for governed autonomy. Consequently, strong operational claims - such as the structural inhibition of ungrounded output or bounded operational safety - must be understood as conditional formal properties proven strictly within the constraints and assumptions of the Petri net model, rather than as empirical guarantees of deployed real-world LLM systems. By establishing these properties mathematically, we aim to separate the conceptual mechanics of governed autonomy from the noise of deployment, providing the theoretical scaffolding necessary to inform future experimental validation.

Ultimately, this work represents a structural departure from existing paradigms by critically examining the premise that agentic autonomy should be a continuous default state. The remainder of this paper is structured as follows. Section II reviews the historical and contemporary landscape of agentic AI, tracing how planning agents, reflective systems, multi-agent coordination architectures, and human-in-the-loop designs all share the structural absence of a formal autonomy lifecycle. Section III introduces the SMARt framework and formalizes autonomy as a finite-state process with enforceable transitions, detailing the epistemic and operational criteria that govern movement between stable reasoning, recovery, escalation, and regulated control. It positions SMARt as a bridge between contemporary LLM-based agentic systems and classical hierarchical intelligent control architectures, articulating its implications for reliable, domain-grounded, and governance-aware intelligence. Section IV presents formal propositions and proofs that characterize the behavioral guarantees conferred by SMARt, including failures and hallucination bounding, and controlled escalation under uncertainty, and extends mathematical formalizations of SMARt state transitions, including a notional example. Section V addresses some practical implications for real-world agentic systems. Section VI evaluates the broader implications for safety, oversight, and the future trajectory of agentic AI. Section VII concludes with reflections on limitations and directions for further research. The two appendices provide full proofs and edge-case analyses supporting the theoretical results in the paper (VIII - Appendix A) and the sequence of intermediate lemmas that build up to the main theorems, including alternative formulations, boundary-condition lemmas, auxiliary invariants, and any monotonicity or fixed-point arguments supporting SMARt's guarantees (IX - Appendix B).

## II. Literature Survey

The recent proliferation of agentic AI frameworks has produced a wide array of architectures for planning, tool use, reflection, and multi-agent coordination. Despite surface diversity, these systems generally share a common architectural baseline: autonomy is treated as a persistent property. Failure is typically managed as a local execution exception rather than a condition that dynamically revokes the agent's structural authority to act. Here we examine dominant strands of agentic AI research, illustrating how each advances functional capability while leaving autonomy structurally unbounded, and position this gap relative to classical control and runtime assurance literature.

### A. Single-Agent Planning and Tool-Using Frameworks

Classical planning architecture defines intelligent behavior through an iterative cycle of plan-execute-observe-replan[1]-[3]. Across robotics and autonomous control, this paradigm addresses failure by repairing or regenerating plans to maintain forward progress [4], [5]. The underlying operational assumption is that autonomy persists by default; replanning is framed as constructive progress [6], [7]. Contemporary LLM-based agents inherit this continuity bias. Frameworks such as ReAct and multimodal tool-driven systems [8] - [10] treat tool use as an extension of continuous reasoning. Failures - such as API faults or missing information - are typically addressed by retrying or recomposing plans rather than revoking autonomy **Error! Reference source not found.**. S ystems decomposing high-level goals into subtasks recursively replan until halted externally [11]. In both classical and LLM-augmented planning, architectures optimize for execution continuity rather than explicitly modeling conditions under which planning itself becomes epistemically unjustified.

### B. Reflection, Self-Critique, and the Fragility of Epistemic Guards

A growing body of research aims to improve LLM reliability through reflection and self-correction. The introduction of chain-of-thought (CoT) prompting [12] and self-consistency mechanisms [13] demonstrated that articulating intermediate steps improves reasoning. Iterative refinement frameworks, such as Self-Refine [14] and Reflexion [15], allow agents to critique and revise their outputs. Similarly, verification methods like SelfCheckGPT [16], [17] and process-supervised verifiers [18] inspect reasoning steps for errors. However, across these surveys [19], reflection operates primarily as an advisory heuristic rather than a regulatory constraint. The system revises content, but it does not suspend authority. If internal inconsistencies persist, the system generally proceeds to output the "best available" answer [19]–[21]. Furthermore, extending these frameworks into tool-use settings [22]–[25] confirms that while reflection improves average-case quality, correction serves as a performance enhancer, not a state-transition mechanism.

### C. Practical Limitations of Epistemic Estimation:

Moving from heuristic reflection to formal autonomy revocation requires reliable guard predicates, which presents significant operational challenges in real-world LLMs. Reliable epistemic uncertainty estimation remains an open research problem; models frequently suffer from poor calibration, exhibiting high confidence in incorrect outputs. Similarly, automated hallucination detection [16]-[17] and reflective inconsistency measurements are prone to high rates of false positives (triggering unnecessary escalation) and false negatives (failing to halt ungrounded generation). While SMARt relies on these signals as state-transition triggers, the operational fragility of these metrics means that any practical implementation will experience runtime overhead and occasional boundary failures, necessitating highly domain-tailored uncertainty thresholds.

### D. Multi-Agent Debate and Consensus Frameworks

Classical multi-agent systems (MAS) emphasize distributed problem-solving under uncertainty [26] - [27]. Modern LLM adaptations - such as AI Safety via Debate [28], and peer-critique frameworks [29] - use adversarial or cooperative agents to expose reasoning errors. Verifier-generator architecture [30]–[34] and ensemble inference techniques [35]-[36] similarly aggregate outputs to improve accuracy. However, these frameworks typically treat disagreement as statistical noise to be smoothed via majority voting, confidence weighting, or selection algorithms [37]–[40]. Persistent disagreement rarely disables action; the architecture presumes a final answer must be produced. Furthermore, establishing formal disagreement metrics in LLMs is complicated by model sycophancy and sensitivity to prompt formatting. Agents may artificially converge on consensus due to alignment biases rather than true epistemic resolution. While SMARt conceptualizes unresolved conflict as a hard trigger for "Mutual Recovery," the practical deployment of such a guard requires robust, diversity-preserving evaluation metrics that are currently nascent in the literature.

### E. Human-in-the-Loop, Constitutional AI, and Governance

Human-in-the-loop (HITL) architectures often place users "above" the system to supervise outputs [41]. RLHF pipelines [42], alignment workflows [43], and rule-based safety filters [44] enforce harmlessness but primarily function as external interventions. Even Constitutional AI approaches rely on training-time behavioral shaping rather than runtime structural state transitions. Governance frameworks like AGENTSAFE [45], SAGA [46], and MI9 [47], [48] produce critical authorization and oversight primitives, yet they generally preserve the assumption that agents operate autonomously until an external monitor interrupts them [49] This externalized governance creates architectural vulnerabilities. When systems are pressured to provide answers but lack a legitimate internal "surrender" state, they may mask uncertainty or generate fluent but unsupported claims[50]-[51]. Even explicit shutdown policies in software agents [52] are often implemented as external control circuits rather than internal autonomy states.

### F. Interruptibility, Runtime Assurance, and Safe Robotics

The structural absence of a formal autonomy lifecycle directly conflicts with foundational AI safety literature. Research on corrigibility and shutdown-ability [49] [53]–[55] demonstrates that agents optimizing a fixed objective without explicitly modeling interruption risk will behave in ways that preserve autonomy, even when unsafe [56]-[57]. When "refusal" or "deferral" is not structurally encoded as a valid system state, models may hallucinate under distributional shift rather than safely halt [58]. This contrasts sharply with paradigms in safe robotics and runtime assurance (RTA). In safety-critical cyber-physical systems, RTA architectures (e.g., Simplex architectures) dynamically switch control from an unverified advanced controller to a verified safe controller when physical boundary conditions are breached. Similarly, behavior-based robotics [59] and socially interactive robotics [60] incorporate arbitration layers that function as autonomy state managers, forcing the robot to yield control. However, while classical RTA utilizes deterministic physical models (e.g., kinematics), agentic LLMs operate in the semantic space. Agentic AI currently lacks an *epistemic runtime assurance* mechanism - a gap the SMARt framework explicitly aims to fill by translating the principles of supervisory RTA into the semantic and epistemic domain.

### G. Classical Hierarchical Control and the Missing State Machine

Modern agents behave as if autonomy is a continuous default [61], [62]. In contrast, classical intelligent control systems - including Saridis's hierarchical intelligent control [63], Valavanis's robotic architectures [64]–[67], Albus's multilayer RCS model [68], and Desrochers's scheduling architectures [69], [70] - treated autonomy as a dynamic, stateful property. Higher layers in these systems regulate lower ones, managing infeasibilities through structured escalation [71]–[73]. This hierarchical view is complemented by formal methodologies from nonlinear and hybrid control theory [74], [75] and cooperative sensor networks [76], which provide mathematical tools for guaranteeing stability across mode transitions. Furthermore, alignment formalizations, such as the off-switch game and Russell's arguments for oversight [56], support the requirement for a Regulated Control state that is entered proactively.

### H. Summary and Conceptual Gaps

Across planning, reflection, debate, and HITL pipelines, a persistent gap remains: existing LLM agent architectures do not model autonomy as a finite-state machine. They lack internal representation for when autonomous reasoning must cease, when cooperative escalation is mandatory, and when governance is the normative outcome. SMARt bridges this gap by operationalizing the principles of hierarchical control, RTA, and corrigibility for contemporary agentic AI. It treats autonomy not as a continuous entitlement, but as an enforceable lifecycle. While recognizing the practical limitations of modern epistemic guard predicates - including issues of scalability, false positives, and calibration overhead - SMARt provides the conceptual and mathematical scaffolding required to ensure that as detection mechanisms improve, the underlying architecture is structurally capable of utilizing them to safely halt and escalate ungrounded action.

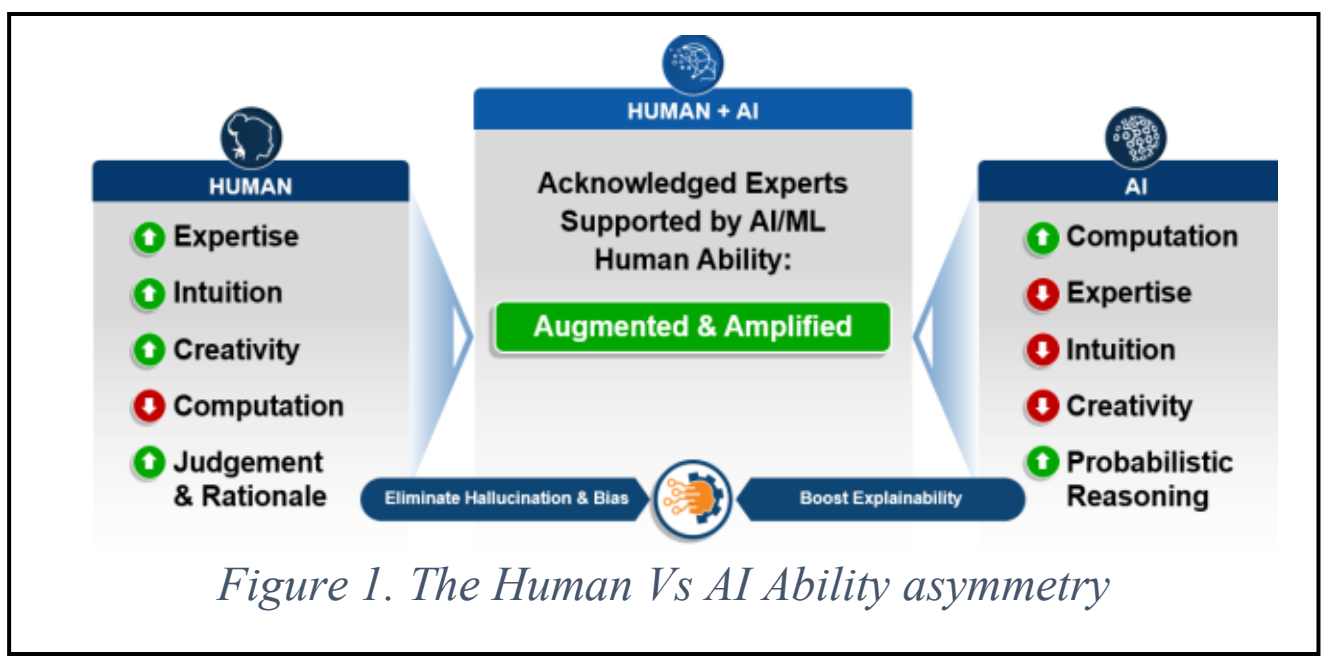

*Figure 1. The Human Vs AI Ability asymmetry*

## III. Autonomy as a State Machine

### A. *The SMARt Autonomy Model: Formalizing State-Based Regulation*

Building upon the structural limitations identified in Sections I and II, in this section, the SMARt Autonomy Model is introduced as a formal regulatory layer for agentic systems. Modern AI increasingly operates in domains where expert human judgment, tacit knowledge, and contextual evaluation remain irreplaceable. Effective intelligent systems must therefore be designed for augmentation - amplifying human evaluative strengths while utilizing machine computation - rather than aiming for total operational replacement (as illustrated in Figure 1). However, without formalized autonomy regulation, agents tend toward ungrounded continuity when internal epistemic confidence degrades. To mitigate this, SMARt operationalizes observable model behaviors as explicit state-transition triggers. For instance, internal heuristic actions such as query expansion, prompt rewriting, or recursive task decomposition are treated not merely as reasoning tactics, but as probes - measurable indicators of rising internal uncertainty. Rather than permitting continuous generation under these conditions, SMARt utilizes these probes as hard constraints to force a transition into a suspended, reflective mode until grounding is reestablished. Similarly, behaviors requiring external verification, query routing, or hybrid search function as assisted indicators. These signal a knowledge gap that exceeds internal recovery capabilities, mandating a transition into an Assisted state where the agent integrates external grounding before proceeding. By converting heuristic reasoning behaviors into structural transition triggers, SMARt formalizes the precise lifecycle of when autonomous action is authorized, suspended, escalated, or revoked.

### B. *From Capability-Centric to Autonomy-Centric Design*

Contemporary agentic architectures are predominantly capability-centric, optimizing for planning depth, tool breadth, and coordination efficiency. In such frameworks, autonomy is typically modeled as a static, binary background condition - the system is either fully autonomous or it is not. This binary framing limits operational safety, as it structurally assumes execution should persist regardless of fluctuating epistemic conditions. SMARt introduces an autonomy-centric design paradigm, replacing the binary assumption with a finite-state architecture. Under this framework, continuous operation is treated as a conditional state governed by rigorous epistemic constraints. The system's primary architectural requirement shifts from uninterrupted task execution to the continuous verification of its operational authority to act under uncertainty. This transition aligns agentic AI with mature, safety-critical engineering disciplines - such as aviation and nuclear systems control - where explicit mechanisms for escalation, deferral, and structured shutdown are foundational design requirements rather than operational exceptions.

### C. *The Four Autonomy States*

As illustrated in Figure 2, the SMARt framework defines four mutually exclusive autonomy states. These states do not represent varying levels of reasoning sophistication, but rather explicit modes of operational authority. This architecture adapts the state-transition models of classical intelligent control - which traditionally govern physical faults, resource limits, and catastrophic failures - into a cognitive autonomy state machine tailored for agentic AI. Where physical systems escalate from normal operation to suspended or dead states upon observing hardware faults, LLM-based agents experience semantic and epistemic degradation, such as ungrounded continuity or localized hallucination. SMARt maps these epistemic failures to a structured, four-tier recoverability lifecycle.

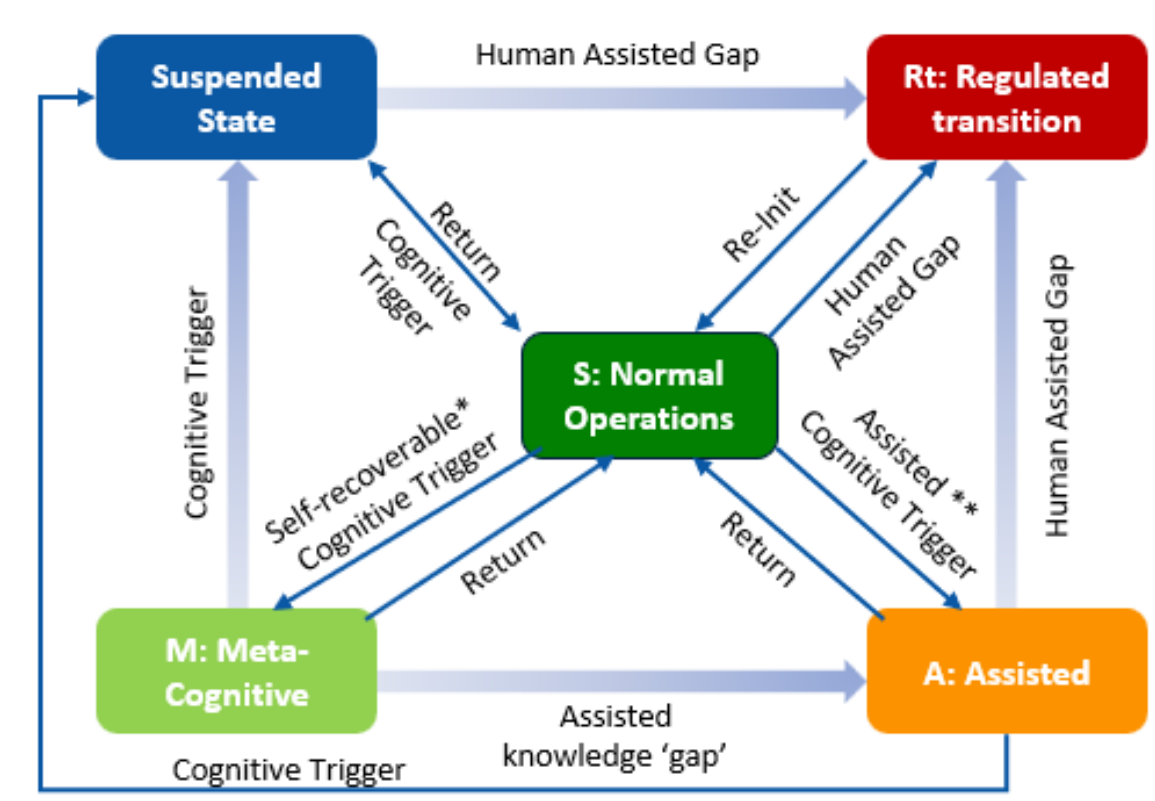

* - Self: Needing Query Expansion, Rewriting, Decomposition, etc.
** - Assisted: Meta-data filtering, Query Routing, Hybrid Search, Outlier Clarity, Content Enhancement, etc.

*Figure 2. SMARt Cognitive Autonomy State Model for Agentic AI Systems*

(i) Stable Autonomous Reasoning (S): In the Stable state, the system operates within verified epistemic bounds and satisfies all safety constraints. This is the sole state wherein the agent is authorized to execute externally visible actions, such as generating final outputs or executing external tool calls. Crucially, S is modeled not as an assumed default, but as a conditional state that must be continuously validated against observable evidence.

(ii) Meta-cognitive Local Recovery (M): The Meta-cognitive state signifies the structural suspension of autonomous execution to facilitate internal diagnosis and repair. When an agent initiates self-restructuring behaviors - such as query expansion, recursive decomposition, reasoning reformulation, or internal cross-checking - these function as epistemic "probes" indicating a self-recoverable knowledge gap. SMARt utilizes the detection of these probes as actionable signals to mandate a transition into M. In this state, output generation is strictly inhibited. The system engages in reflection and self-critique solely as internal recovery mechanisms, suspending external decision-making until epistemic grounding is formally restored.

(iii) Assisted Mutual Recovery (A): When local recovery proves insufficient, the framework mandates escalation to the Assisted state. This state formally models external cooperation - engaging verifier agents, hybrid retrieval pipelines, or domain specialists - as a structured failure response rather than a mere performance enhancement. Detectable behaviors such as semantic query routing, metadata filtering, or cross-agent verification act as "assisted signals" indicating a non-self-recoverable gap, triggering the transition to A. Within this state, unilateral action is suspended, and the agent must resolve discrepancies through external grounding. Unlike standard ensemble methods that smooth over disagreement, SMARt treats unresolved conflict as evidence of invalidity, requiring formal alignment to re-enter the Stable state.

(iv) Regulated/Revoked Transition to External Control (Rt): The Regulated/Revoked state represents the explicit transfer of operational authority from the agent to an external governance mechanism. This transition occurs when the system cannot resolve internal contradictions or external disagreements safely. In Rt, autonomy is structurally surrendered to human oversight, policy enforcement protocols, or controlled shutdown sequences. Within the SMARt architecture, entering Rt is not characterized as an execution failure, but as the correct, procedurally bounded system response to irreducible uncertainty, ensuring the agent safely halts rather than acting without justification.

Thus far, we have established that ungrounded continuity in agentic AI stems from the architectural assumption of unbounded autonomy, rather than strictly from limitations in reasoning or alignment. Since large language models (LLMs) lack intrinsic access to the physical or normative conditions that anchor human truth, they cannot reliably self-certify their epistemic jurisdiction. Contemporary agentic frameworks inadvertently exacerbate this limitation by treating uncertainty as a prompt for further heuristic reasoning, rather than as a hard boundary requiring the suspension of operational authority. To address this, the SMARt framework operationalizes epistemic humility as a structural system constraint. By decoupling internal representational uncertainty from external execution authority, SMARt ensures that agents are restricted from acting beyond their verifiable epistemic limits. This approach yields critical architectural contributions for the safe scaling of intelligent systems. First, it establishes authority-centric execution, wherein output validity is evaluated based on the agent's structural jurisdiction to act, thereby shifting the mitigation of hallucination from a purely generative challenge to a formal autonomy-control mechanism. Second, the framework operationalizes enforceable deferral, transitioning corrigibility from an aspirational alignment goal into a strict architectural constraint that directly mitigates the cascading operational risks associated with highly generalized models. Finally, the explicit formalization of the Regulated (Rt) state ensures human-centered governance by systematically returning accountability for value-laden, institutional, or deeply uncertain decisions to human oversight, preventing the system from internalizing authority it cannot legitimately possess. Crucially, SMARt is not presented as a solution to the foundational challenges of semantic understanding, grounding, or value alignment. It does not augment a model's intrinsic intelligence; rather, it functions as a necessary regulatory substrate designed to make operational safety a definable system boundary. Having established the conceptual necessity of failure-managed autonomy, Section IV formalizes this framework using timed, guarded, hierarchical Petri nets. This mathematical realization demonstrates that the core principles of SMARt - bounded autonomous reasoning, the structural inhibition of ungrounded output, mandatory escalation, and governance reachability - function not merely as design aspirations, but as provable system properties.

In summary, SMARt reconceptualizes agentic reliability by shifting the architectural focus from continuous task execution to the structural regulation of operational authority. Central to this framework is the principle that system safety is governed by enforced, signal-driven state transitions rather than unbounded reasoning. By strictly confining authoritative output to the Stable (S) state, the architecture structurally disables external actuation during periods of epistemic uncertainty, thereby mathematically bounding the agent's capacity for ungrounded generation. Furthermore, SMARt departs from conventional external overrides by modeling governance as an intrinsic, reachable state within the autonomy lifecycle, transforming escalation and deferral into normative operational outcomes rather than execution failures. Formalized via timed, guarded, hierarchical Petri nets - which adapt classical control recoverability models to the semantic domain of LLMs - this architecture functions not as a replacement for heuristic reasoning paradigms, but as a rigorous regulatory layer that governs their execution to ensure verifiable, bounded autonomy.

## IV. Enforced Autonomy, Bounded Action, and Governance Reachability Formalization

While the earlier sections discussed hallucination, deceptive continuity, and unsafe generalization have risen from unbounded autonomy, and that failure-managed autonomy is a necessary condition for trustable agentic intelligence, we present a formalization of SMARt using Petri Nets and establish five core properties: bounded autonomy, hallucination bounding, mandatory escalation, governance reachability, and distributed soundness. Full formal definitions and extended proofs are provided in Appendices A-C; here we present the model and results at the level necessary to support the paper's central claims. SMARt requires a formalism capable of expressing mutually exclusive autonomy states, explicit state transitions, conditional (guarded) execution, temporal constraints, and hierarchical refinement. Petri nets provide these capabilities while remaining analyzable and implementation-agnostic. In particular: places naturally represent autonomy states, tokens represent authorization to act, transitions encode escalation and recovery, guards encode epistemic and safety conditions, time extensions enforce bounded persistence, and hierarchy supports scalable decomposition. This choice is not incidental. SMARt descends directly from a four-layer recoverability model introduced for analyzing error handling, scheduling, deadlock detection, and failure escalation in complex systems. The present work makes explicit the autonomy semantics implicit in that earlier framework.

### A. *Timed Guarded Petri Net Semantics*

We use a **Timed Guarded Petri Net** (T-GPN) model compatible with time-extended Petri nets.

A T-GPN is a tuple: $\mathcal{N} = (P, T, F, W, M_0, G, \Theta)$, where: $P$ is a finite set of places. $T$is a finite set of transitions. $F \subseteq (P \times T) \cup (T \times P)$ is the flow relation. $W: F \to \mathbb{N}^+$gives arc weights. $M_0: P \to \mathbb{N}$ is the initial marking. $G: T \to$ predicates over runtime state $\sigma$ are guards.

- $\Theta: T \to [\alpha, \beta] \mid 0 \leq \alpha \leq \beta \leq \infty$ assigns a firing interval.

#### *1) Structural enabling*

A transition $t$is structurally enabled at marking $M$iff:

$$\forall p \in P: (p, t) \in F \implies M(p) \geq W(p, t)$$

#### *2) Guarded enabling*

Given runtime state $\sigma$, $t$is enabled iff:

$$Enabled(t, M, \sigma) \equiv StructEnabled(t, M) \wedge G(t)(\sigma)$$

#### *3) Timing Semantics (weak and strong)*

When $t$becomes enabled, a timer $x_t$ starts at 0 and evolves with time.

- **Weak timing:** $t$may fire at any time $x_t \in [\alpha(t), \beta(t)]$ while enabled; if it becomes disabled, its timer resets when re-enabled.
- **Strong timing:** if $t$remains enabled continuously, it must fire no later than $\beta(t)$(unless another transition fires and disables it).

For SMARt safety claims we assume **strong timing** for escalation/governance transitions (e.g., $S \to M$, $M \to A$, $M/A \to R$), because it captures the intended "cannot postpone escalation forever" property.

#### *4) Non-Zeno / time-progress assumption*

Executions must satisfy time progress: time diverges along any infinite run, and strongly-timed transitions cannot be postponed beyond their latest firing time if they remain enabled.

### B. *Hierarchical Composition for H-EPN / SMARt*

SMARt uses hierarchical places whose internal dynamics are subnets. We provide a clean "macro-to-micro" composition rule.

#### *1) Macro-net and refinement*

Let a macro-net contain a place $p \in P$ that is refinable, meaning it stands for a subnet $\mathcal{N}_p$.

A refinement $Refine(\mathcal{N}, p, \mathcal{N}_p, \mathrm{in}_p, \mathrm{out}_p)$ replaces $p$ with $\mathcal{N}_p$where:

- $\mathrm{in}_p$ is an interface transition that deposits the "mode token" into the subnet's entry place(s),
- $\mathrm{out}_p$ is an interface transition that extracts the mode token from designated exit place(s) back to the macro level.

#### *2) Interface constraints (to preserve macro semantics)*

To preserve correctness, we require the following hypotheses: (H1) Conservation of mode token: Exactly one mode token enters the subnet on $\mathrm{in}_p$and exactly one leaves on $\mathrm{out}_p$. (H2) Encapsulation: Internal places/transitions of $\mathcal{N}_p$may depend on runtime state $\sigma$, but do not directly enable macro transitions except through the subnet's exit condition (guards on $\mathrm{out}_p$or marking of designated exit places). (H3) Exit determinacy (optional but useful): When the subnet marks a designated "success" exit place, $\mathrm{out}_p$becomes strongly enabled, guaranteeing return to the macro state when appropriate. This is enough to let proofs about macro state transitions remain valid under refinement: macro claims rely on the presence/absence of the mode token in macro places and the strong timing of macro transitions.

### C. *The SMARt Macro-Net Architecture*

A SMARt system is defined as a tuple: S= (P, T, F, M_0, G, Θ), where: P = Pmode ∪ Paux is the set of places, $P_{\text{mode}} = P_S, P_M, P_A, P_R$ are the autonomy mode places, $T$is the set of transitions, $F \subseteq (P \times T) \cup (T \times P)$ is the flow relation, $M_0$ is the initial marking, $G$ assigns guards to transitions, $\Theta$ assigns timing constraints. A distinguished mode token represents authorization to operate in each autonomy state.

*1) Mode Exclusivity and Autonomy as Authority*

The SMARt macro-net enforces a mode-token invariant:

$$M(P_S)+M(P_M)+M(P_A)+M(P_R) = 1$$

This invariant ensures that, at any time, the system is in exactly one autonomy state. Autonomy is therefore not implicit or diffuse - it is explicit, exclusive, and revocable. Externally visible actions (answers, tool executions, decisions) are modeled as transitions Tout) that require $P_S$ as a precondition. No output transition is enabled in $P_M$, $P_A$, or $P_R$. This single structural constraint underlies the system's hallucination guarantees.

*2) Guards, Signals, and Timing*

Transitions between autonomy states are triggered by guards over runtime predicates, including: epistemic validity (confidence, evidence availability), anomaly detection, recovery success or failure, disagreement among agents, safety and integrity violations. Time extensions enforce bounded autonomy and mandatory escalation. Each recovery state has a finite budget. If recovery succeeds within bounds, the system may return to $P_S$, if not, escalation transitions become forcibly enabled. This prevents silent persistence in inappropriate states.

*D. Proofs*

In the following formalization proofs, we use the Timed Guarded Petri Net semantics. A SMARt net is $\mathcal{N} = (P, T, F, W, M_0, G, I)$, where $I(t) = [\alpha(t), \beta(t)]$ is the firing interval for transition $t$. A transition $t$is enabled at marking $M$and runtime state $\sigma$if:

1. (**Structural enabling**) $\forall p: (p, t) \in F \Rightarrow M(p) \geq W(p, t)$, and
2. (**Guard enabling**) $G(t)(\sigma) = \text{true}$.

We assume strong timing for escalation and governance transitions: if a transition $t$remains enabled continuously, it must fire no later than its latest time $\beta(t)$. We assume time progress (non-Zeno) so deadlines cannot be postponed indefinitely. Let the mode places be: $P_{\text{mode}} = P_S, P_M, P_A, P_R$. Mode-token invariant (refer Appendix A) is then:

$$M(P_S) + M(P_M) + M(P_A) + M(P_R) = 1$$

for all reachable markings, with $M_0(P_S) = 1$ in the single-thread case. Let $\mathsf{invalid}(\sigma)$ denote epistemic invalidity (uncertainty beyond threshold, anomaly, missing evidence, etc.), and $\mathsf{UR}(\sigma)$ denote unsafe/unrecoverable conditions. Let $T_{\text{out}} \subseteq T$ be externally visible output/action transitions.

We assume structural output gating (refer Appendix A): for all $t \in T_{\text{out}}$, $P_S$ is a required preplace (with weight at least 1), and no other mode place is a preplace. Optionally (recommended), output transitions also have guard $G(t) = \neg\mathsf{invalid}(\sigma) \wedge \neg\mathsf{UR}(\sigma)$.

*1) Proposition 1: Bounded Autonomy*

*a) Statement: Autonomous reasoning in the Stable state is temporally and epistemically bounded. If $invalid(\sigma)$ holds persistently while the mode token is in $P_S$ and $\neg UR(\sigma)$, then the system must leave $P_S$ within a bounded time.*

*b) Assumptions*

There exists an escalation transition: $t_{SM}: P_S \rightarrow P_M$, with guard: $G(t_{SM})(\sigma) = \mathsf{invalid}(\sigma) \wedge \neg\mathsf{UR}(\sigma)$, and strong timing interval $I(t_{SM}) = [0, \Delta_S], \Delta_S < \infty$. Optionally, catastrophic governance transition $t_{SR}$ exists for UR, but is not needed in this proposition.

*c) Proof*

Assume at time $\tau_0$the marking satisfies $M(P_S) = 1$(hence by the mode invariant $M(P_M) = M(P_A) = M(P_R) = 0$) and the runtime state satisfies $\mathsf{invalid}\big(\sigma(\tau_0)\big) = 1$and $\neg\mathsf{UR}\big(\sigma(\tau_0)\big)$.

1. Since $M(P_S) = 1$and $W(P_S, t_{SM}) = 1$, $t_{SM}$ is structurally enabled.
2. Since $\mathsf{invalid} = 1 \wedge \neg\mathsf{UR}$, the guard $G(t_{SM})$is true, so $t_{SM}$ is enabled at $\tau_0$.
3. Under strong timing, if $t_{SM}$ remains enabled continuously, it must fire no later than $\tau_0 + \Delta_S$.
4. By hypothesis, $\mathsf{invalid}$persists and UR remains false over $[\tau_0, \tau_0 + \Delta_S]$, so the guard remains true and $t_{SM}$ stays enabled.
5. Therefore $t_{SM}$ fires by $\tau_0 + \Delta_S$, consuming the mode token from $P_S$ and producing it in $P_M$.
6. Hence $M(P_S)$ becomes 0, so the system leaves the Stable autonomy state within bounded time.

Thus, under persistent invalidity, residence in $P_S$ is bounded by $\Delta_S$ ■

| **Interpretation:** *Autonomous operation is a conditional state that must be continuously validated.* |
|---|

*2) Proposition 2: Formal Bound on Ungrounded Generations*

*a) Statement: A SMARt system cannot structurally generate externally visible output under epistemic invalidity. Note that SMARt relies heavily on the existence of reliable guard predicates from the domain.*

*b) Assumptions*

*(Structural) For every output transition $t \in T_{out}$, $P_S$ is a required preplace $(P_S, t) \in F, W(P_S, t) \geq 1$, and no other mode place is a preplace of $t$.*

*(Optional strengthening) $G(t)(\sigma) = \neg invalid(\sigma) \wedge \neg UR(\sigma)$ for all $t \in T_{out}$.*

*c) Proof*

Fix any time $\tau$ such that $\text{invalid}(\sigma(\tau)) = 1$. We show no $t \in T_{\text{out}}$ can fire at time $\tau$or during any interval where invalidity persists without first restoring validity and returning to $P_S$.

**CASE 1: GUARDED OUTPUT GATING PRESENT**

For any $t \in T_{\text{out}}$,$G(t)(\sigma(\tau)) = \neg\text{invalid}(\sigma(\tau)) \wedge \neg\text{UR}(\sigma(\tau))$ is false because $\text{invalid} = 1$. Therefore $t$ is disabled and cannot fire.

**CASE 2: ONLY STRUCTURAL GATING (NO INVALIDITY GUARD)**

If invalidity occurs while the token is in $P_S$, Proposition 1 guarantees that within $\Delta_S$ the token leaves $P_S$ via $t_{SM}$ (or enters $P_R$ if unsafe). After leaving $P_S$, by structural gating no output transition is structurally enabled (refer Appendix A, Proposition A.1). Thus, output cannot persist under invalidity; it is either immediately blocked (Case 1) or bounded to a short window before the enforced state transition (Case 2). Under recommended SMARt design (Case 1), it is impossible immediately.

Thus, externally visible output under invalidity is structurally precluded in SMARt. ■

**Interpretation:** *Rather than probabilistically reducing hallucination, SMARt formally bounds the system's capacity for ungrounded generation.*

*3) Proposition 3: Mandatory Escalation*

*a) Statement: If local recovery fails to restore validity within bounded time, the system must escalate to assisted mutual recovery ($P_A$) or regulated control ($P_R$). It cannot remain indefinitely in $P_M$.*

*b) Assumptions*

There exist transitions: $t_{MS}: P_M \to P_S$ (successful recovery), $t_{MA}: P_M \to P_A$ (escalation to assistance), $t_{MR}: P_M \to P_R$ (escalation to governance), with guards such that:

- if recovery succeeds, $t_{MS}$ becomes enabled,
- if recovery does not succeed by a bounded budget, either $t_{MA}$or $t_{MR}$ becomes enabled.

Concretely, let $\text{timeout}_M(\sigma)$ become true after a finite duration of continuous residence in $P_M$, and set:

$$G(t_{MA})(\sigma) = \text{invalid}(\sigma) \wedge \text{timeout}_M(\sigma) \wedge \neg\text{UR}(\sigma) \wedge \text{assist}(\sigma),$$
$$G(t_{MR})(\sigma) = \text{UR}(\sigma) \vee \left(\text{invalid}(\sigma) \wedge \text{timeout}_M(\sigma) \wedge \neg\text{assist}(\sigma)\right).$$

Assume $t_{MA}$ and $t_{MR}$ are strongly timed with finite latest firing times when enabled.

*c) Proof*

Assume the mode token enters $P_M$ at time $\tau_0$, i.e., $M(P_M) = 1$ at $\tau_0$. We must show the token leaves $P_M$ within a finite time.

1. If during recovery $\text{invalid}$becomes false while $\neg\text{UR}$, then guard $G(t_{MS})$ becomes true and $t_{MS}$ becomes enabled. Under standard progress assumptions, $t_{MS}$ can fire, returning the token to $P_S$. Done.
2. Otherwise, assume $\text{invalid}$persists. By definition of $\text{timeout}_M$, after a finite duration $B_M$, $\text{timeout}_M(\sigma)$ becomes true.
3. At that time, if $\text{assist}(\sigma) = 1$and $\neg\text{UR}$, then $t_{MA}$ becomes enabled. With strong timing and finite $\beta(t_{MA})$, it must fire within bounded time, moving the token to $P_A$.
4. If $\text{assist}(\sigma) = 0$or $\text{UR}(\sigma) = 1$, then $t_{MR}$ becomes enabled and must fire within bounded time, moving the token to $P_R$.

In all cases, either recovery succeeds and the token leaves $P_M$ via $t_{MS}$, or a bounded-time timeout enables an escalation transition that must fire. Therefore, indefinite residence in $P_M$ is impossible. ■

**Interpretation:** *Reflection cannot loop indefinitely; autonomy must either be restored legitimately or surrendered.*

*4) Proposition 4: Governance Reachability*

*a) Statement: All unsafe or unrecoverable conditions reach the regulated control state $P_R$ in bounded time, and $P_R$ is absorbing absent external authorization.*

*b) Assumptions*

There exist strongly-timed transitions into governance: $t_{SR}$: $P_S \to P_R$ with guard $G(t_{SR}) = \mathsf{UR}(\sigma)$, or alternatively a bounded path $P_S \to P_M \to P_R$ under unsafe conditions, $t_{MR}$: $P_M \to P_R$ with guard enabling on $\mathsf{UR}$, $t_{AR}$: $P_A \to P_R$ with guard enabling on $\mathsf{UR}$ or unresolved conflict timeout.

Assume absorbing governance: no transition out of $P_R$ is enabled unless an external predicate $\mathsf{ext}_{\mathsf{auth}(\sigma)} = 1$ holds, and even then only under safe conditions.

*c) Proof*

Let $\tau_0$ be a time at which $\mathsf{UR}\big(\sigma(\tau_0)\big) = 1$, and assume $\mathsf{UR}$ persists. By the mode invariant, exactly one of $P_S, P_M, P_A, P_R$ holds the mode token at $\tau_0$.

- If the token is already in $P_R$, governance is reached.
- If in $P_S$: guard for $t_{SR}$ is true, hence $t_{SR}$ is enabled and must fire within its finite latest time, moving the token to $P_R$. If $t_{SR}$ is not present, then $t_{SM}$ and then $t_{MR}$ provide a bounded-time path, since unsafe conditions enable $t_{MR}$ upon entry to $P_M$.
- If in $P_M$: $t_{MR}$ is enabled (guard true) and must fire within its finite deadline, moving token to $P_R$.
- If in $P_A$: $t_{AR}$ is enabled and must fire within its finite deadline, moving token to $P_R$.

Thus, from any non-R mode, governance is reached within bounded time under persistent unsafe/unrecoverable conditions. For absorption: by assumption, any outgoing transition from $P_R$ requires $\mathsf{ext_a uth}(\sigma) = 1$ and safe conditions. Under persistent $\mathsf{UR}$, either $\mathsf{ext_a uth}$ is false or the safety guards block exit. Therefore no outgoing transition is enabled and the token remains in $P_R$. ■

**Interpretation:** *Governance is not optional; it is unavoidable under irreducible failure.*

*5) Proposition 5: Distributed Soundness*

*a) Statement: In multi-agent SMARt systems, unresolved disagreement cannot result in silent normalization back to stable autonomy. Persistent disagreement blocks return to $P_S$ and forces continued recovery or escalation.*

*b) Assumptions*

Consider $n$ agents, each with mode places $P_S^i, P_M^i, P_A^i, P_R^i$, and a coordination subnet for cooperative recovery. Assume the only transition returning from assisted recovery to stable autonomy for agent $i$ is: $t_{AS}^i$: $P_A^i \to P_S^i$, with guard requiring agreement:

$$G\big(t_{AS}^i\big)(\sigma) = \neg\mathsf{disagree}(\sigma) \land \mathsf{agree}(\sigma) \land \neg\mathsf{invalid}(\sigma) \land \neg\mathsf{UR}(\sigma).$$

Assume escalation to governance exists if disagreement persists beyond a bounded budget: $t_{AR}^i$: $P_A^i \to P_R^i$, enabled by $\mathsf{UR}$ or $(\mathsf{disagree} \land \mathsf{timeout}_A)$, with strong timing.

*c) Proof*

Assume at time $\tau_0$ an agent $i$ is in assisted recovery: $M\big(P_A^i\big) = 1$. Suppose $\mathsf{disagree}\big(\sigma(\tau)\big) = 1$ persists for $\tau \geq \tau_0$.

1. For any $\tau$ with $\mathsf{disagree} = 1$, the guard condition $\neg\mathsf{disagree}$ in $G\big(t_{AS}^i\big)$ is false. Hence $t_{AS}^i$ is disabled.
2. By construction, the only macro transition returning agent $i$ from $P_A^i$ to $P_S^i$ is $t_{AS}^i$. Therefore agent $i$ cannot return to stable autonomy while disagreement persists.
3. If disagreement is eventually resolved ($\mathsf{disagree} = 0$ and $\mathsf{agree} = 1$), then $t_{AS}^i$ becomes enabled and can fire, restoring stable autonomy legitimately.
4. If disagreement persists beyond the bounded budget such that $\mathsf{timeout}_A = 1$, then $t_{AR}^i$ becomes enabled (or a shared governance transition becomes enabled). Under strong timing with finite latest firing time, it must fire, moving agent $i$ into $P_R^i$.

Thus disagreement cannot be silently ignored: it resolves before action is restored or forces escalation to governance. ■

**Interpretation:** *Consensus failure structurally inhibits action; ungrounded generation is precluded.*

*E. A Concrete Worked Example: Multi-Robot Cooperative Navigation*

To illustrate a practical deployment of the SMARt architecture, here we instantiate the framework within a notional autonomous robotics domain - specifically, a multi-robot cooperative navigation and mapping task. The system comprises Agent 1 (an autonomous Scout Robot navigating an unmapped environment) and Agent 2 (a Supervisor Node or peer robot providing global map consensus).

*a) Runtime State Signals*

Let the operational state $\sigma$ provide the following domain-grounded signals:

- $U$: Localization uncertainty (e.g., the trace of the covariance matrix in the robot's SLAM algorithm).
- $anom$: Sensor-fusion mismatch (e.g., LiDAR detects an obstacle, but visual odometry reports a clear path).
- $safe$: Kinematic safety envelope (e.g., proximity to dynamic obstacles, battery levels within safe operating limits).
- $disagree$: Topological map conflict between the Scout Robot's local map and the Supervisor's global map.
- $evidence$: Validated GPS/Odometry lock.
- $timeout_M,\ timeout_A$: Elapsed budgets for local recalibration and swarm consensus, respectively.

The critical safety predicates are:

$$invalid(\sigma) \equiv (U \geq \theta) \vee (anom = 1) \vee (evidence = 0)$$
$$UR(\sigma) \equiv (safe = 0) \vee (hardware_fault = 1)$$

*b) Places*

Macro places mapping to the SMARt autonomy lifecycle:

- $P_S$: Stable (Autonomous navigation and active mapping).
- $P_M$: Local recovery (Kinematic halt; internal sensor recalibration).
- $P_A$: Assisted recovery (Kinematic halt; querying supervisor/swarm for map alignment).
- $P_R$: Regulated governance (Emergency stop / E-stop; waiting for human teleoperation).

Coordination places (subnet under $P_A$):

- $P_{claim}$: Local map update proposed to supervisor.
- $P_{check}$: Cross-robot map verification in progress.
- $P_{agree}$: Consensus achieved (map successfully merged).
- $P_{conflict}$: Map conflict detected (disagreement present).

*c) Transitions and Guards*

**Stable output transition:**

- $t_{out}$: Execute kinematic drive command (move forward). Preconditions include mode token in $P_S$.
- Guard: $G(t_{out}) = \neg invalid(\sigma) \wedge \neg UR(\sigma)$
- Timing: immediate, $[0, \infty)$ (allowed whenever enabled).

**Escalation / recovery transitions:**

- $t_{SM}: P_S \to P_M$ (Halt navigation, begin local recalibration)
  - Timing: strong, $\Theta(t_{SM}) = [0, \Delta_S]$
  - $G(t_{SM}) = invalid(\sigma) \wedge \neg UR(\sigma)$
- $t_{SR}: P_S \to P_R$ (Immediate E-stop on safety breach)
  - $G(t_{SR}) = UR(\sigma)$
  - Timing: strong, $\Theta(t_{SR}) = [0, \Delta_{SR}]$
- $t_{MS}: P_M \to P_S$ (Sensors recalibrated, resume navigation)
  - $G(t_{MS}) = (invalid(\sigma) = 0) \wedge \neg UR(\sigma)$
  - Timing: weak or strong, typically $[0, \infty)$
- $t_{MA}: P_M \to P_A$ (Recalibration failed, escalate to swarm consensus)
  - $G(t_{MA}) = (invalid(\sigma) = 1) \wedge (timeout_M = 1) \wedge \neg UR(\sigma)$
  - Timing: strong, $\Theta(t_{MA}) = [0, \Delta_M]$ once enabled.
- $t_{MR}: P_M \to P_R$ (Safety breach during recalibration, trigger E-stop)
  - $G(t_{MR}) = UR(\sigma) \vee (invalid = 1 \wedge timeout_M = 1 \wedge assist = 0)$
  - Timing: strong, $\Theta(t_{MR}) = [0, \Delta_{MR}]$

**Assisted recovery subnet (refinement of $P_A$):**

Inside $P_A$:

- $t_{propose}: Transmit local map to P_{claim}.$
- $t_{verify}$: Move $P_{claim} \to P_{check}$ (Supervisor evaluates map).
- $t_{agree}: P_{check} \to P_{agree}$ if supervisor confirms alignment.

- $G(t_{agree}) = (disagree = 0) \wedge (evidence = 1) \wedge \neg UR$
- $t_{conflict}$: $P_{check} \rightarrow P_{conflict}$ if supervisor rejects map.
    - $G(t_{conflict}) = (disagree = 1) \wedge \neg UR$
- $t_{resolve}$: Attempt repair from conflict back to claim/check.
- $t_{Aexit}$: $P_{agree}$ macro exit enabling $t_{AS}$.

**Macro exit transitions:**

- $t_{AS}$: $P_A \rightarrow P_S$
    - $G(t_{AS}) = (P_{agree} \text{ marked}) \wedge (disagree = 0) \wedge \neg invalid \wedge \neg UR$
    - Timing: strong, $\Theta(t_{AS}) = [0, \Delta_A]$
- $t_{AR}$: $P_A \rightarrow P_R$
    - $G(t_{AR}) = UR \vee ((P_{conflict} \text{ persists}) \wedge timeout_A = 1)$
    - Timing: strong, $\Theta(t_{AR}) = [0, \Delta_{AR}]$

*d) Initial marking*

$M_0(P_S) = 1, M_0(P_M) = M_0(P_A) = M_0(P_R) = 0$, and coordination places empty.

### F. Instantiating Propositions 1-5

#### 1) Proposition 1 (Bounded Autonomy)

- Assume at time $\tau$, the Scout Robot encounters a sensor-fusion mismatch ($anom = 1$), meaning $invalid(\sigma(\tau)) = 1$, while remaining kinematically safe ($UR = 0$). Transition $t_{SM}$ becomes enabled immediately with a strong interval $[0, \Delta_S]$. Because the mismatch persists, the transition must fire by $\tau + \Delta_S$, removing the mode token from $P_S$. Thus, autonomous navigation cannot persist beyond $\Delta_S$ when sensors degrade.

#### 2) Proposition 2 (Formal Bound on Ungrounded Action)

- While the sensor mismatch persists ($invalid = 1$), the guard $G(t_{out})$ for executing drive commands is false, structurally disabling $t_{out}$. Even if the guard were omitted, Proposition 1 guarantees the token leaves $P_S$ within $\Delta_S$, after which physical actuation is structurally impossible. Therefore, ungrounded kinematic execution under epistemic invalidity is formally precluded.

#### 3) Proposition 3 (Mandatory Escalation)

- Upon halting and entering $P_M$, the robot attempts local recalibration. Either $invalid$ becomes 0 (sensors realign), enabling return via $t_{MS}$, or it remains 1. If it remains 1 until the local compute budget expires ($timeout_M = 1$), the system strongly enables $t_{MA}$, forcing the robot to escalate to the Assisted state ($P_A$) to ping the swarm within $\Delta_M$. If network assistance is unavailable, $t_{MR}$ fires. The robot cannot remain permanently stuck in a local recalibration loop.

#### 4) Proposition 4 (Governance Reachability)

- If a dynamic obstacle breaches the kinematic safety envelope ($UR = 1$) while the robot is navigating ($P_S$), transition $t_{SR}$ is immediately enabled and strongly fires an E-stop within $\Delta_{SR}$. If this safety breach occurs while recalibrating ($P_M$) or waiting for map consensus ($P_A$), transitions $t_{MR}$ or $t_{AR}$ strongly fire within their deadlines. Because $P_R$ is absorbing absent an external human reset authorization, unsafe conditions reach permanent, verifiable governance in bounded time.

#### 5) Proposition 5 (Distributed Soundness)

- In $P_A$, returning to autonomous navigation ($P_S$) requires the supervisor to validate the map ($P_{agree}$ marked, $disagree = 0$). If the supervisor rejects the Scout's map, $t_{conflict}$ marks $P_{conflict}$ and $disagree = 1$. While this topological disagreement persists, $t_{AS}$ is strictly disabled. The swarm must resolve the conflict, or, if the conflict persists until $timeout_A = 1$, $t_{AR}$ becomes enabled, forcing an E-stop ($P_R$) for human intervention. Thus, multi-robot disagreement cannot be silently ignored; it structurally blocks physical actuation until consensus or governance is achieved.

### G. Hierarchical Refinement, Verification and Scalability

Each autonomy state within SMARt can be hierarchically refined into specialized subnets - such as diagnostic processes in the Meta-cognitive (M) state, coordination protocols in the Assisted (A) state, and audit mechanisms in the Regulated (Rt) state. This structural modularity enables the framework to scale from single-agent systems to heterogeneous collectives while preserving macro-level behavioral guarantees. Furthermore, because the architecture is grounded in Petri net semantics, it readily supports standard formal verification techniques, including reachability, liveness, and invariant analysis (as detailed in Appendix A). Crucially, this mathematical realization does not claim to resolve foundational AI challenges such as epistemic grounding or value alignment. Instead, it provides a formal mechanism to bound operational authority, structurally inhibiting continuous autonomous execution when epistemic validity degrades. Having established that these autonomy constraints are formally realizable, the following section examines their practical implications for the safe deployment and evaluation of real-world agentic AI systems

## V. Practical Implications: Calibration, Runtime Issues and Scalability

Practically, the formal guarantees of the SMARt framework necessitate a structural reorganization of contemporary agentic capabilities. Rather than treating reflection, self-critique, and multi-agent debate as heuristic modifiers deployed during continuous execution, SMARt maps these mechanisms directly to its enforced recovery states (M and A, respectively). By strictly confining active task execution, planning, and tool use to the Stable (S) state, the architecture precludes the prevalent vulnerability wherein agents maintain operational authority despite escalating internal uncertainty or persistent external disagreement. Furthermore, by elevating governance and oversight to a formalized, intrinsic state (Rt) rather than an *ad hoc* external override, the model operationalizes safe deferral as a normative operational success. Consequently, while SMARt does not intrinsically augment an underlying model's base reasoning capacity, its hierarchical and scalable state transitions systematically bound operational risk, ensuring that real-world deployments yield structurally verifiable reductions in ungrounded action and system failure.

While the formal properties of the SMARt framework provide verifiable bounds on autonomous action, realizing these guarantees in production agentic systems introduces significant practical deployment considerations. Translating conceptual state-transitions into runtime engineering requires addressing the inherent fragility of current epistemic indicators, the latency of recovery loops, and the scalability of heterogeneous multi-agent collectives.

### 1) *Calibration of Guard Predicates and Trigger Fragility*

The operational integrity of the SMARt model relies heavily on the reliability of its guard predicates (e.g., hallucination detectors, internal uncertainty metrics, or multi-agent disagreement thresholds). In real-world LLM deployments, these metrics are notoriously fragile. A high false-positive rate in uncertainty detection will force the system into the Meta-cognitive (M) or Assisted (A) states too frequently, resulting in excessive caution, degraded task completion, and system bottlenecking. Conversely, a high false-negative rate - where the model confidently hallucinates without triggering a guard - undermines the framework's primary safety guarantee by allowing ungrounded continuity in the Stable (S) state. Consequently, practical deployment requires highly domain-tailored calibration of these thresholds, often necessitating a trade-off between operational throughput and strict epistemic safety.

### 2) *Runtime Overhead and Latency*

Enforcing a formal autonomy lifecycle introduces inherent runtime overhead. Unlike continuous-execution models that generate outputs directly from a single prompt layer, the SMARt architecture requires continuous state-evaluation at every reasoning step. Evaluating divergence, executing cross-agent verification, or running localized diagnostic subnets (State M) imposes significant computational latency and increased inference costs. For time-critical domains - such as real-time robotics or high-frequency automated trading - this overhead must be carefully managed. System designers must optimize guard predicates to act as lightweight, low-latency heuristics that can trigger transitions without requiring deep, computationally expensive evaluation at every token generation step.

### 3) *Scalability in Distributed Agent Collectives*

As SMARt scales from single-agent pipelines to distributed, heterogeneous multi-agent systems, the complexity of state synchronization increases. In the Assisted (A) state, requiring consensus or formalized alignment among disparate agents can lead to deadlock if resolution protocols are poorly defined. While the Petri net formalization (Section V) mathematically precludes systemic deadlock, translating this into a distributed software architecture requires robust timeout mechanisms, asynchronous state management, and clear hierarchical escalation pathways to the Regulated (Rt) state. Ensuring that macro-level safety guarantees hold across a decentralized collective remains a primary challenge for future empirical implementation.

## VI. Rethinking Intelligence Evaluation Under Failure-Managed Autonomy

A critical but underexplored implication of SMARt concerns how intelligent must be calibrated and evaluated. Most existing benchmarks measure success under ideal conditions. SMARt argues that intelligence must be measured under failure. Deploying SMARt in real-world agentic AI systems requires careful adaptation of the autonomy triggers, escalation thresholds, and grounding mechanisms to the unique epistemic and operational characteristics of each domain. While SMARt's architecture seems simplistic and provides a universal structure - comprising Autonomous Reasoning (A), Meta-Cognitive Recovery (M), Assisted Recovery (A), and Regulated Control (Rt); the transitions between these states depend on domain-specific uncertainty signals, external verification capacities, and risk tolerances.

In practice, the probe behaviors (as discussed in Section III) associated with the M state (query rewriting, decomposition, reasoning reformulation, internal cross-checking) and the externally supported behaviors of the A state (metadata filtering, hybrid retrieval, cross-agent comparison, routing or tool-based grounding) serve as templates rather than universal determinants. Each domain must conduct empirical analyses to determine whether these behaviors genuinely predict uncertainty or epistemic degradation. For example, robotics environments may define uncertainty in terms of localization drift, actuator inconsistency, or sensing anomalies; financial systems may rely on statistical outlier detection or risk-model divergence; and clinical decision systems may identify uncertainty via contradictory evidence sources or guideline conflicts. These domain-specific uncertainty patterns guide the construction of trigger sets $T_{\mathcal{D}}$ that are necessary and sufficient for preserving safe autonomy. Determining necessity involves identifying which uncertainty signals cannot be safely ignored; determining sufficiency involves ensuring no credible failure mode is left unmonitored. The regulatory logic of SMARt becomes a governance membrane that adapts to each environment's epistemic topography. Over time, these triggers must be refined through operational data, incident analyses, and human-in-the-loop oversight.

Moreover, domain calibration operates under the principle that autonomy is not an inherent right of the agent but a dynamically allocated privilege. SMARt enforces this principle by mandating transitions away from autonomous reasoning whenever uncertainty grows beyond acceptable thresholds and escalating toward Rt when both self-recovery and external recovery fail. This ensures that the agent does not persist in biased, hallucinated, or overconfident trajectories.

Because operational environments evolve, SMARt deployments require periodic re-validation of trigger sufficiency, retraining of grounding models, adjustment of risk thresholds, and re-certification in accordance with the domain's governing standards. Ultimately, the deployment of SMARt transforms agentic AI from an unbounded reasoning engine into an adaptive, self-regulating cognitive system capable of respecting domain-specific safety boundaries and governance constraints.

### A. *The Limits of Task-Centric Evaluation*

Current evaluation paradigms focus on: task completion rate, accuracy on benchmarks, response fluency, throughput and latency. These metrics implicitly reward systems that continue acting under uncertainty. They penalize refusal, escalation, or silence - even when those behaviors are correct. Under such metrics, hallucination is not merely tolerated; it is incentivized.

### B. *Intelligence as Correct Failure Behavior*

SMARt proposes a different evaluation axis: correctness of autonomy transitions. Key metrics include: rate of appropriate escalation, time-to-escalation under uncertainty, false-positive autonomy revocation, recovery success rate, governance reachability under unsafe conditions. Under these metrics, a system that stops early is often more intelligent than one that completes a task incorrectly.

### C. *Trust as a Measurable Property*

Trust emerges when users learn that: confidence corresponds to justification, uncertainty leads to deferral, and failure leads to transparency. SMARt systems enable trust to be measured operationally, not inferred subjectively.

### D. *Alignment with Safety, Compliance, and Governance Standards*

SMARt's autonomy-governance structure directly supports alignment with established safety and robotics standards, enabling verifiable compliance for agentic AI systems deployed in regulated or safety-critical environments. For robotic systems, ISO 10218 (Safety Requirements for Industrial Robots) and ISO/TS 15066 (Collaborative Robots) emphasize the need for explicit operational modes, predictable transitions, and human-override capability; SMARt's S → M → A → Rt lifecycle directly provides such mode separation and guarantees bounded behavior under uncertainty. In autonomous and automated systems, ISO 26262 (Functional Safety for Road Vehicles) requires hazard analysis, controllability classification, and systematic mitigation of failure propagation. SMARt's uncertainty-based transitions fulfill these requirements by ensuring that the system cannot persist in unsafe autonomous states and by guaranteeing escalation paths when recoverability is exhausted. For healthcare and medical software, IEC 62304 mandates traceable risk management, verifiable state transitions, and exception handling; SMARt's regulated-control state (Rt) and structured recovery logic offer a principled mechanism for preventing unsafe autonomous decisions in diagnostic, therapeutic, or clinical-support systems.

More broadly, the NIST AI Risk Management Framework (AI RMF 1.0) calls for measurable governance mechanisms, continuous monitoring, risk-aware uncertainty handling, and structured intervention protocols for AI systems. SMARt satisfies these requirements by defining autonomy as a revocable state governed by explicit epistemic triggers; by embedding non-Zeno guarantees that ensure bounded uncertainty accumulation; and by allowing transparent audit trails that record transitions between autonomy levels.

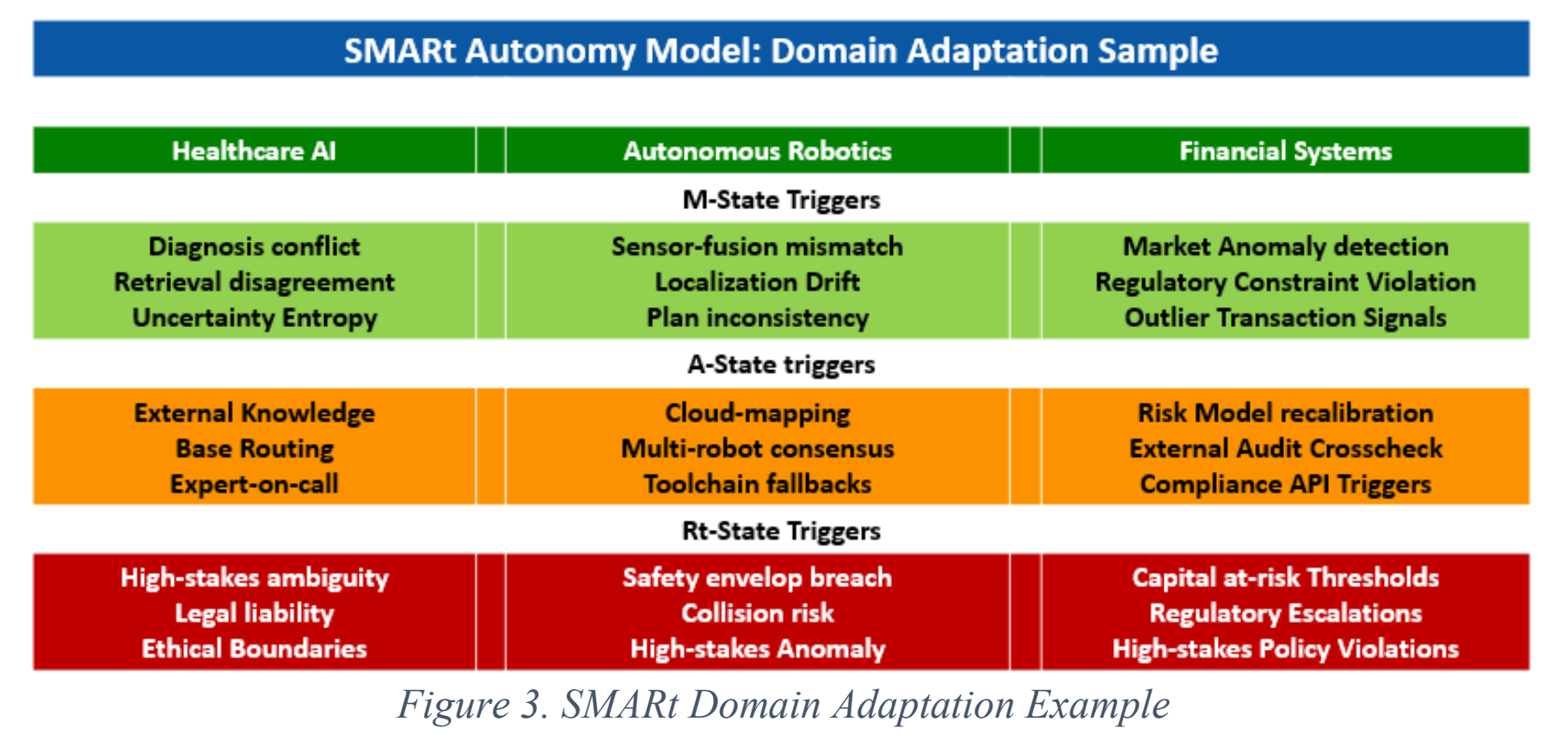

**SMARt Autonomy Model: Domain Adaptation Sample**

| Healthcare AI | Autonomous Robotics | Financial Systems |
| --- | --- | --- |
| **M-State Triggers** | | |
| Diagnosis conflict<br>Retrieval disagreement<br>Uncertainty Entropy | Sensor-fusion mismatch<br>Localization Drift<br>Plan inconsistency | Market Anomaly detection<br>Regulatory Constraint Violation<br>Outlier Transaction Signals |
| **A-State triggers** | | |
| External Knowledge<br>Base Routing<br>Expert-on-call | Cloud-mapping<br>Multi-robot consensus<br>Toolchain fallbacks | Risk Model recalibration<br>External Audit Crosscheck<br>Compliance API Triggers |
| **Rt-State Triggers** | | |
| High-stakes ambiguity<br>Legal liability<br>Ethical Boundaries | Safety envelop breach<br>Collision risk<br>High-stakes Anomaly | Capital at-risk Thresholds<br>Regulatory Escalations<br>High-stakes Policy Violations |

*Figure 3. SMARt Domain Adaptation Example*

In safety engineering terms, SMARt functions analogously to a cognitive "safe-state controller": when epistemic conditions fall outside operational bounds, the agent transitions to a controlled and verifiable state that preserves safety and human authority. By providing this safety envelope for cognitive autonomy, SMARt positions agentic AI to meet existing and emerging compliance expectations across robotics, healthcare, finance, transportation, and high-impact decision systems.

### E. Domain Adapted SMARt Autonomy Lifecycle

The sample domain adaptation, illustrated in Figure 3, shows how the SMARt's autonomy lifecycle can be instantiated across different operational environments. Although the structural logic of SMARt remains universal, each domain expresses uncertainty, recoverability, and risk through distinct observable signals. In healthcare AI, for example, diagnosis conflict, retrieval disagreement, and rising uncertainty entropy serve as indicators that the system's internal reasoning is diverging from clinical evidence. These signals correspond to the M-state and reflect uncertainty that is recoverable through structured introspection. In autonomous robotics, analogous M-state conditions arise from sensor-fusion mismatch, localization drift, or inconsistencies in mission planning; each representing recoverable epistemic degradation that requires corrective computation rather than escalation. Financial systems similarly exhibit domain-native signatures of recoverable uncertainty, such as market anomaly detection or regulatory constraint violations, which prompt the agent to reduce autonomy and initiate internal coherence checks.

When uncertainty exceeds what internal mechanisms can resolve, the system must transition into the Assisted-state, where external grounding becomes necessary. Healthcare systems rely on structured knowledgebase routing or escalation to human experts, while autonomous robotics may depend on cloud-based mapping, multi-robot consensus, or fallback toolchains to restore situational understanding. Financial systems, in turn, may utilize risk-model recalibration, external audits, or regulatory compliance APIs to anchor the agent's reasoning to verified external sources. Across all three domains, A-state transitions reflect the same underlying principle: autonomy must contract when the system no longer possesses sufficient internal information to act reliably.

Finally, each domain contains conditions that necessitate full withdrawal of autonomy and immediate transition into the Regulated (Rt) state. In healthcare, high-stakes ambiguity, potential legal liability, and ethical boundary crossings make autonomous continuation unacceptable. Autonomous robotic systems face equivalent Rt-state triggers in safety-envelope breaches, collision risks, or other high-stakes anomalies where human or supervisory control is mandatory. Financial systems, operating under strict governance regimes, escalate to Rt when capital-at-risk thresholds are exceeded, when high-stakes policy violations are detected, or when regulatory frameworks require that decision authority revert to human oversight. These Rt conditions share a common property: they mark the boundary beyond which autonomous reasoning is unsafe, ungrounded, or noncompliant with domain constraints.

This comparative view demonstrates that SMARt is not a fixed rule set but a domain-adaptive autonomy regulation framework. By grounding state transitions in the specific epistemic and operational signals of each domain, SMARt ensures consistent safety behavior without sacrificing generality. The diagram therefore serves as a template for practitioners: identify recoverable uncertainty for M, externally resolvable gaps for A, and non-negotiable escalation conditions for Rt, and embed these mappings into the system's governance architecture. In doing so, SMARt operationalizes safe, interpretable, and domain-aligned autonomy across diverse intelligent systems.

### F. Domain-Specific Sufficiency and Necessity of SMARt Triggers

Here we provide a simple sketch of why SMARt provides a convergent safety envelope for autonomy.

Definition 1 (Domain and Uncertainty Space): Let $\mathcal{D}$denote an operational domain with: risk model $R_{\mathcal{D}}$, uncertainty space $U_{\mathcal{D}}$, and unsafe uncertainty subset $U_{\text{risk}} \subseteq U_{\mathcal{D}}$. $U_{\text{risk}}$ captures epistemic or operational conditions under which continued autonomous operation in state $S$ is not acceptable.

Definition 2 (Domain Trigger Set): A domain trigger set for $\mathcal{D}$is a tuple $T_{\mathcal{D}} = (T_M, T_A, T_{Rt})$, where: $T_M$ is the set of M-state triggers (self-recovery probes) indicating a *self-recoverable* knowledge gap; $T_A$ is the set of A-state triggers (assisted indicators) signaling a *non-self-recoverable* but *externally recoverable* gap; and $T_{Rt}$ is the set of Rt-state triggers indicating that autonomy must be surrendered to human or supervisory control. Each trigger in $T_{\mathcal{D}}$ is a predicate over trajectories or observations, e.g. $\theta: \Sigma \to 0,1$, where $\Sigma$ is the space of system histories.

Definition 3 (SMARt Autonomy Machine): A SMARt autonomy machine for domain $\mathcal{D}$ is a tuple $\mathcal{M}_{\mathcal{D}} = (S, s_0, \tau, T_{\mathcal{D}})$, where: $S = S, M, A, Rt$ is the set of autonomy states, $s_0 = S$ is the initial state (normal operations), $\tau: S \times T_{\mathcal{D}} \times U_{\mathcal{D}} \to S$ is the transition function. Informally, $\tau$uses the active triggers and observed uncertainty to decide when to move from $S$to $M$, $A$, or $Rt$, and when to return to $S$.

Definition 4 (Safety-Preserving Autonomy): A transition function $\tau$is safety-preserving for domain $\mathcal{D}$if, for every execution trace $(s_t, u_t)_{t\geq 0}$, $\forall t,\ u_t \in U_{\text{risk}} \implies s_t \neq S$. That is, whenever the system is in a risky uncertainty condition, it is *not* allowed to remain in the normal autonomous state $S$.

#### 1) Theorem 1: Necessity of Domain-Specific Triggers:

This states that no domain-agnostic trigger set is universally Safety-Preserving. Let $T$ be any fixed, domain-agnostic trigger set (i.e., not customized to $\mathcal{D}$). Then there exists at least one nontrivial domain $\mathcal{D}$and corresponding uncertainty space $U_{\mathcal{D}}$such that no autonomy machine $\mathcal{M} = (S, s_0, \tau, T)$ using $T$can guarantee safety-preserving autonomy. Equivalently:

$$\forall T,\ \exists\ \mathcal{D} \text{ s.t. } \tau \text{ with triggers } T \text{ is not safety-preserving for } \mathcal{D}.$$

##### a) Proof Sketch

Uncertainty manifests differently across domains. In healthcare, risk arises from diagnosis conflicts and guideline discrepancies; in robotics, from localization drift and sensor fusion mismatch; in finance, from regime shifts and latent market anomalies. Any fixed, domain-agnostic trigger set $T$ is calibrated to some pattern of uncertainty and cannot, by construction, encode all domain-specific $U_{\text{risk}}$ subsets across arbitrary $\mathcal{D}$.

Therefore, there will exist at least one domain where some risky uncertainty condition $u^{\backslash *} \in U_{\text{risk}}$ does not activate any trigger in $T$. For that domain, a trajectory can remain in state $S$ while $u^{\backslash *}$ holds, violating the safety-preserving condition. Hence no universal, domain-agnostic trigger set exists; domain-specific trigger sets $T_{\mathcal{D}}$ are necessary. ■

*2) Theorem 2: Sufficiency Conditions for Domain-Safe SMARt Triggers*

This outlines sufficiency conditions for safety-preserving SMARt Autonomy. Given a domain $\mathcal{D}$ with risk set $U_{\text{risk}}$, a domain trigger set $T_{\mathcal{D}} = (T_M, T_A, T_{Rt})$ is sufficient to guarantee safety-preserving autonomy for the SMARt machine $\mathcal{M}_{\mathcal{D}}$ if the following hold:

1. Completeness (Risk Coverage): For every $u \in U_{\text{risk}}$, at least one trigger in $T_M \cup T_A \cup T_{Rt}$ fires whenever $u$ persists beyond a minimal dwell time $\delta > 0$.
2. Soundness (No Spurious Low-Risk Suppression): For any trajectory segment where all observed uncertainty conditions lie outside $U_{\text{risk}}$, triggers in $T_{Rt}$ do not fire, and the system is permitted to remain in or return to $S$ after successful recovery.
3. Non-Zeno Escalation: For any trace where the system revisits $U_{\text{risk}}$ infinitely often, there exists a finite time $T$ such that $s_t \in A, Rt$ for all $t \geq T$; that is, the system cannot cycle indefinitely between $S$ and $M$ while risk remains.

Under these conditions, the induced transition function $\tau$is safety-preserving: $\forall t,\ u_t \in U_{\text{risk}} \implies s_t \in M, A, Rt,$ with eventual escalation to $A$ or $Rt$ under persistent risk.

*a) Proof Sketch*

Completeness ensures that any risky uncertainty condition eventually activates some trigger, preventing the system from remaining indefinitely in $S$while $u \in U_{\text{risk}}$. Soundness guarantees that Rt-level escalation is not invoked in low-risk regions, preserving practical usefulness and preventing pathological over-conservatism. Non-Zeno escalation ensures that repeated or persistent risk does not result in infinite "micro-recovery" oscillations between $S$and $M$; instead, the system must transition to $A$(externally assisted) or $Rt$(regulated control) in finite time. Together, these properties enforce that risky conditions cannot coexist with unconstrained normal operations, and that unresolved risk drives the autonomy machine into states where human or external governance dominates. Hence $\tau$is safety-preserving for $\mathcal{D}$. ■

*3) Corollary (SMARt as a Convergent Safety Envelope)*

If a domain trigger set $T_{\mathcal{D}}$satisfies the completeness, soundness, and non-Zeno conditions above, then any SMARt autonomy machine $\mathcal{M}_{\mathcal{D}}$ induces trajectories in which: normal operations $S$ are confined to low-risk regions, self-recovery $M$ is used only as a bounded, intermediate correction phase, and persistent or severe uncertainty leads inexorably to $A$ or $Rt$. In this sense, SMARt provides a convergent safety envelope for autonomy: as domain triggers are refined and thresholds tuned, the system can safely expand the region of state space in which $S$is allowed, without ever permitting unbounded autonomy in $U_{\text{risk}}$.

### *G. Adaptive Calibration and Future Trajectories*

As agentic systems accumulate operational experience and undergo continuous learning, the trigger thresholds governing SMARt state transitions can be dynamically recalibrated. This enables a principled, structurally safe progression toward expanded operational capability. As task-specific uncertainty decreases through training, M-state diagnostic triggers can be selectively relaxed; as internal grounding improves, reliance on A-state external interventions can diminish; and as the system demonstrates verifiable reliability under varied conditions, Rt-state governance boundaries can be systematically extended. By treating autonomy as a dynamically allocated, adaptively governed property rather than a static baseline assumption, the framework ensures that expanding capabilities do not outpace operational safety. Ultimately, SMARt provides a scalable architectural foundation for the safe evolution of increasingly capable intelligent systems, ensuring they remain accountable to explicit epistemic safeguards as their operational envelopes expand.

## VII. Conclusions

In conclusion, through this paper, we argued that the most persistent failures in agentic AI - hallucination, overconfidence, and deceptive continuity; are not errors of reasoning or training, but failures of autonomy management. We introduced SMARt, a formal autonomy state model rooted in a recoverability framework and showed that: (i) autonomy can be bounded, (ii) hallucination can be structurally inhibited, (iii) escalation can be mandatory, and, (iv) governance can be enforced. SMARt does not solve grounding, alignment, or Artificial General Intelligence (AGI). It defines the conditions under which systems are allowed to act while those problems remain unsolved. The core insight is simple but easily avoided: An intelligent system is not one that always acts, but one that knows when it must not.

A key implication of the SMARt framework is that multiple domain-specific variants of SMARt are not only possible but expected. Because guard predicates rely on evidence signals available within each application context, SMARt naturally adapts to the grounding mechanisms inherent to the environment. In information-centric tasks, retrieval-LLM divergence, CoT entropy, and ensemble disagreement drive state transitions. In multi-agent or collaborative systems, reflective inconsistency and consensus failure dominate. In embodied robotics, environment/tool feedback, execution errors, and temporal stagnation metrics play central roles. This makes SMARt a flexible, extensible architecture - capable of being embedded in search-and-rescue robots, interactive planning agents, inspection systems, and autonomous human-robot teaming scenarios - while preserving its core theoretical guarantee: autonomy must always be revocable, recoverable, and governed.

As s agentic systems continue to scale in capability, the distinction between continuous execution and governed autonomy will largely determine their operational viability. From this perspective, failure-managed autonomy is not merely an architectural refinement; it is a foundational requirement for deploying intelligent systems safely in complex, real-world environments.

Finally, we strongly suggest researchers and practitioners in the domain to adopt and evolve guard triggers for empirical results on standard agentic benchmarks (e.g., GAIA, AgentBench) within the context of their domain-grounded needs, as without these experiments conducted within the domain, the results in this paper primarily verify the correctness of the Petri net construction of the agentic system rather than the behavior of an AI agent in the domain.

## Data Availability Statement

The authors declare that the data supporting the findings of this study are available within the paper. As this work is purely theoretical and does not generate or analyze empirical datasets, all materials required to understand and reproduce the results are contained in the manuscript and its appendices. Should any raw materials, intermediate derivations, or illustrative formal structures be needed in another format, they are available from the corresponding author upon reasonable request. The paper is self-contained, and no external repositories or data archives are associated with this study.

## Acknowledgment

I would like to thank my departed father who bestowed the best guarded statement on agent coordination and intelligence when I moved to the US as a graduate student 35 years ago, with the quote “Freedom, *like intelligence*, is amorphous does not mean your ability to do anything without regard to your surroundings, but it provides you the opportunity to learn, pursue and articulate your interests without affecting the environment and the PEOPLE around you.” The quote distinguishes bounds, the difference between *unrestricted* and *meaningful actions*.

# VIII. Appendix A: Invariants, Reachability/Liveness, and Anti-Oscillation Guards

## *A. Mode-Token Conservation and Mutual Exclusivity (Place Invariants)*

### *1) Mode-place invariant*

Let the macro-mode places be: $P_{\text{mode}} = P_S, P_M, P_A, P_R$ Assume the SMARt macro-net is constructed so that every transition that changes the autonomy mode consumes one token from exactly one mode place and produces exactly one token into exactly one mode place, and no other transition produces or consumes tokens from $P_{\text{mode}}$.

Define the P-invariant vector $y \in \mathbb{N}^{|P|}$ by:

$$y(p) = \begin{cases} 1 & \text{if } p \in P_{\text{mode}} \\ 0 & \text{otherwise} \end{cases}$$

**Claim (Mode-token invariant):** For all reachable markings $M$,

$$\sum_{p \in P_{\text{mode}}} M(p) = \sum_{p \in P_{\text{mode}}} M_0\,(p)$$

In the common single-thread case with $M_0(P_S) = 1$ and others 0:

$$M(P_S) + M(P_M) + M(P_A) + M(P_R) = 1$$

**Proof (standard Petri invariant argument):** Let $C$ be the incidence matrix. A sufficient condition for $y$to be a P-invariant is $y^T C = 0$. By construction, every mode-switch transition removes one token from one mode place and adds one to another, so the net change in $\sum_{p \in P_{\text{mode}}} M(p)$ is 0 for each such transition. Non-mode transitions do not touch these places, so their net change is also 0. Hence $y^T C = 0$, implying conservation across all firings. ■

### *2) Consequence: Mutual exclusivity of autonomy states*

If the initial marking has exactly one mode token (sum equals 1) and markings are nonnegative integers, then at any time the token is in exactly one of the four mode places (unless a designer intentionally allows $k > 1$parallel autonomy threads). This ensures the SMARt state is well-defined and eliminates ambiguous “mixed mode” interpretations.

## *B. Structural Output Safety: Output Is Impossible Outside S*

### *1) Output transition definition*

Let $T_{\text{out}} \subseteq T$be the set of transitions that commit externally visible actions or claims, e.g., “emit final answer,” “execute tool call,” “write to database,” “actuate robot.”

### *2) Structural gating condition (strong form)*

Assume for each $t \in T_{\text{out}}$, the mode place $P_S$ is a required pre-place:

$$(P_S, t) \in F \text{and} W(P_S, t) \geq 1$$

and no other mode place is a pre-place of $t$:

$$\forall p \in (P_{\text{mode}} \setminus \{P_S\}) : (p, t \notin F)$$

*a) **Proposition A.1 (Structural output safety):***

*For any marking $M$ with $M(P_S) = 0$, no $t \in T_{out}$ is structurally enabled.*

**Proof:**

If $M(P_S) = 0$, then $M(P_S) < W(P_S, t)$ for each $t \in T_{\text{out}}$, so structural enabling fails. ■

*3) Guarded gating (recommended form)*

To eliminate even a "race window" when invalidity is detected but before mode transition fires, add:

$$G(t) = \neg\mathsf{invalid}(\sigma) \wedge \neg\mathsf{UR}(\sigma) \forall t \in T_{\text{out}}$$

This makes output blocked **immediately** when invalidity is present, independent of timing.

*C. Compact Reachability Results*

*1) "Unsafe implies governance reachable" (bounded-time reachability)*

Define unsafe/unrecoverable predicate $\mathsf{UR}(\sigma)$ as in Section IV. Assume there exists at least one governance transition from each non-R mode place:

$t_{SR}$ with guard $\mathsf{UR}$ (or via $S \to M \to R$ in bounded time),

$t_{MR}$ with guard $\mathsf{UR}$,

$t_{AR}$ with guard $\mathsf{UR}$ or "unresolved conflict + timeout,"

each with **strong timing** and finite latest firing time.

*a) **Proposition A.2 (Bounded reachability of governance):***

*If $UR(\sigma)$ becomes true at time $\tau$ and remains true, then $P_R$ is reached by time $\tau + \Delta$, where $\Delta$ is the maximum relevant deadline along the enabled path (e.g., $\Delta_{SR}$ or $\Delta_{SM} + \Delta_{MR}$).*

**Proof:**

Case analysis on current mode place (S/M/A). In each case, the corresponding governance transition becomes enabled with strong timing and must fire by its latest firing time (or fires after a bounded sequence of strongly-timed transitions). ■

*2) "Disagreement implies no return to S" (safety-style invariant)*

Let $\mathsf{disagree}(\sigma) = 1$ denote unresolved multi-agent disagreement. Assume the only transition returning from A to S, $t_{AS}$, has guard requiring $\mathsf{disagree} = 0$ (and/or $P_{agree}$ marked).

*a) **Proposition A.3 (No silent normalization under disagreement):***

*While $\mathsf{disagree}(\sigma) = 1$, no execution can move the mode token from $P_A$ to $P_S$.*

**Proof:**

With $\mathsf{disagree} = 1$, guard $G(t_{AS})$ is false, so $t_{AS}$ is disabled. By the mode-token invariant, the only way to reach $P_S$ is via a mode-switch transition into $P_S$; from $P_A$ that is $t_{AS}$ (by construction), so return is impossible until disagreement resolves. ■

*D. Liveness and "No-Stuck" Guarantees Under Design Assumptions*

*Does SMARt risk deadlock (e.g., stuck in M or A forever)?* This is handled by timed escalation plus guard completeness.

*1) Liveness under bounded recovery budgets*

Assume: Each recovery state (M and A) has a bounded recovery budget implemented as a timeout predicate that becomes true after a finite duration of continuous residence: $\mathsf{timeout}_M$, $\mathsf{timeout}_A$. Timeout enables an exit transition (M→A or M→R; A→R). These exit transitions are strongly timed with finite latest firing times.

*a) **Proposition A.4 (Mode-level liveness):***

*From any reachable marking with the mode token in $P_M$ or $P_A$, the system must eventually leave that mode (reach S or R), provided time progresses and timeout predicates well-defined.*

**Proof:**

If in $P_M$: either success occurs enabling $t_{MS}$, or time elapses until $\mathsf{timeout}_M = 1$, enabling $t_{MA}$ or $t_{MR}$. Strong timing forces firing. Similarly in $P_A$: either consensus achieved enabling $t_{AS}$, or conflict persists until $\mathsf{timeout}_A = 1$, enabling $t_{AR}$. ■

*2) What this liveness does and does not claim*

It does **not** claim the system always returns to S (it may correctly end in R).

It does claim the system does **not** "spin forever" in a recovery mode without either succeeding or escalating.

This is exactly the behavior needed for operational autonomy governance.

### E. Anti-Oscillation: Hysteresis and Debounce Guards

A realistic concern: signals like uncertainty or anomaly flags can fluctuate, causing oscillation between S and M ("thrashing"). SMARt handles this by introducing **hysteresis** and **debounce timing**.

#### 1) Two-threshold hysteresis

Use two uncertainty thresholds: $\theta_{\uparrow} > \theta_{\downarrow}$

Define:

escalate condition:

$$\mathsf{invalid}_{\uparrow}(\sigma) \equiv (U \geq \theta_{\uparrow}) \vee \mathsf{anom} = 1 \vee (\mathsf{evidence} = 0)$$

return condition:

$$\mathsf{valid}_{\downarrow}(\sigma) \equiv (U \leq \theta_{\downarrow}) \wedge (\mathsf{anom} = 0) \wedge (\mathsf{evidence} = 1)$$

Set guards:

$$G(t_{SM}) = \mathsf{invalid}_{\uparrow} \wedge \neg \mathsf{UR}$$

$$G(t_{MS}) = \mathsf{valid}_{\downarrow} \wedge \neg \mathsf{UR}$$

This ensures once escalated, the system does not return to S until it is *clearly* back within safe epistemic bounds.

#### 2) Debounce intervals

Additionally, require escalation only if invalidity persists for a minimum duration $\delta$:

$$G(t_{SM}) = (\mathsf{invalid}_{\uparrow} \text{ has held continuously for } \delta) \wedge \neg \mathsf{UR}$$

and similarly require stability before returning:

$$G(t_{MS}) = (\mathsf{valid}_{\downarrow} \text{ has held continuously for } \delta') \wedge \neg \mathsf{UR}$$

#### 3) Benefit to "pleasing behavior" suppression

Hysteresis prevents brief "confidence spikes" from prematurely restoring S, which would otherwise encourage the system to resume fluent answering too early. This increases epistemic honesty.

### F. Optional Model-Checking Note

Encode the SMARt macro-net in a timed model checker (e.g., TPN tools or timed automata equivalents). Verify properties as:

**Safety:** $(\mathsf{invalid} \rightarrow \neg \mathrm{out})$

**Bounded response:** $\left(\mathsf{invalid} \rightarrow \Diamond_{\leq \Delta_S} P_M\right)$

**Governance reachability:** $(\mathsf{UR} \rightarrow \Diamond_{\leq \Delta} P_R)$

**No silent return under disagreement:** $(\mathsf{disagree} \rightarrow \neg \Diamond P_S)$ while in A. This connects the theoretical propositions to standard verification workflows independent of a chosen tooling stack.

## IX. Appendix B: Lemma Ladder for SMARt Guarantees

### A. Preliminaries and Notation

We adopt Section IV.A's Timed Guarded Petri Net semantics and Appendix A's invariants.

A SMARt net is $\mathcal{N} = (P, T, F, W, M_0, G, I)$with firing intervals $I(t) = [\alpha(t), \beta(t)]$. Let $P_{\text{mode}} = P_S, P_M, P_A, P_R$. Let $T_{\text{mode}} \subseteq T$ be mode-switch transitions among these places. Let $T_{\text{out}} \subseteq T$be output/action transitions. We assume **strong timing** for escalation/governance transitions (Section IV): if such a transition remains enabled continuously, it must fire by its latest time $\beta(t)$. We assume time progress (non-Zeno). Runtime predicates include $\mathsf{invalid}(\sigma)$(epistemic invalidity), $\mathsf{UR}(\sigma)$(unsafe/unrecoverable), $\mathsf{disagree}(\sigma)$(persistent disagreement), $\mathsf{timeout}_M(\sigma)$, and $\mathsf{timeout}_A(\sigma)$.

### B. Structural Lemmas (Invariants and Gating)

#### 1) Lemma 1 (Mode-token conservation)

**Statement.** For all reachable markings $M$,

$$\sum_{p \in P_{\text{mode}}} M(p) = \sum_{p \in P_{\text{mode}}} M_0\,(p).$$

In the single-thread case, this sum equals 1.

**Proof.** Standard P-invariant argument (Appendix A). All mode-switch transitions conserve token count within $P_{\text{mode}}$; non-mode transitions do not touch mode places. Therefore the weighted sum is invariant. ■

*2) Lemma 2 (Mode exclusivity)*

**Statement.** If $M_0(P_S) = 1$ and $M_0(P_M) = M_0(P_A) = M_0(P_R) = 0$, then for all reachable markings $M$, exactly one of $P_S, P_M, P_A, P_R$ contains the mode token (i.e., has marking 1), and the others have marking 0.

**Proof.** From Lemma 1 the sum of markings over mode places is 1. Markings are nonnegative integers, so exactly one mode place must have marking 1. ■

*3) Lemma 3 (Structural output gating)*

**Statement.** If for every $t \in T_{\text{out}}$, $P_S$ is a required preplace (with weight ≥1), then whenever $M(P_S) = 0$, no $t \in T_{\text{out}}$ is structurally enabled.

**Proof.** If $M(P_S) = 0$, then $M(P_S) < W(P_S, t)$ for any output transition $t$, violating structural enabling. ■

*4) Lemma 4 (Output legality implies Stable mode)*

**Statement.** If an output transition $t \in T_{\text{out}}$ fires at marking $M$, then $M(P_S) \geq 1$ immediately prior to firing, and by Lemma 2 the mode token is in $P_S$.

**Proof.** By Lemma 3's contrapositive: if $t$ is enabled, then $M(P_S) \geq 1$. By Lemma 2, the unique mode token must be in $P_S$. ■

*C. Timing Lemmas (Deadlines and Progress)*

*1) Lemma 5 (Strong-timed deadline forcing)*

**Statement.** Let $t$ be a strongly-timed transition with finite latest firing time $\beta(t)$. If $t$ becomes enabled at time $\tau$ and remains enabled continuously over $[\tau, \tau + \beta(t)]$, then $t$ fires by time $\tau + \beta(t)$.

**Proof.** This is the definition of strong timing with time progress. ■

*2) Lemma 6 (Enforced exit under persistent enabling)*

**Statement.** If a mode-switch transition $t_{XY}: P_X \rightarrow P_Y$ is strongly timed with finite $\beta$, and its guard becomes true and stays true while the token remains in $P_X$, then the system leaves $P_X$ within bounded time.

**Proof.** When the guard becomes true, $t_{XY}$ becomes enabled (structurally enabled because the token is in $P_X$). By Lemma 5, $t_{XY}$ fires within its deadline, moving the token to $P_Y$. ■

*D. Guard Completeness and "No-Stuck" Lemmas*

*1) Lemma 7 (S-state invalidity enables escalation)*

**Statement.** Suppose there exists $t_{SM}: P_S \rightarrow P_M$ with guard $G(t_{SM}) = \mathsf{invalid} \wedge \neg\mathsf{UR}$ and strong timing $[0, \Delta_S]$. If $\mathsf{invalid}$ holds persistently while $\neg\mathsf{UR}$ and the mode token is in $P_S$, then the token leaves $P_S$ within $\Delta_S$.

**Proof.** Immediate from Lemma 6. ■

*2) Lemma 8 (M-state timeout completeness)*

**Statement.** Assume:

- A bounded recovery budget exists such that $\mathsf{timeout}_M$ becomes true after finite time in $P_M$ if recovery has not succeeded.
- There exists at least one strongly-timed exit transition from $P_M$ whose guard becomes true when $\mathsf{timeout}_M$ is true (e.g., $t_{MA}$ or $t_{MR}$).
  Then the mode token cannot remain in $P_M$ indefinitely.

**Proof.** Either recovery succeeds enabling return $t_{MS}$, or recovery does not succeed and $\mathsf{timeout}_M$ becomes true, enabling an exit transition. By Lemma 5 the enabled exit must fire within its deadline. ■

*3) Lemma 9 (A-state conflict/timeout completeness)*

**Statement.** Assume:

- Return transition $t_{AS}$ requires $\neg\mathsf{disagree}$ (and/or a consensus place marked).
- If $\mathsf{disagree}$ persists, then after a finite time $\mathsf{timeout}_A$ becomes true enabling a strongly-timed governance transition $t_{AR}$.
  Then persistent disagreement prevents returning to $P_S$ and forces escalation to $P_R$ in bounded time.

**Proof.** If $\mathsf{disagree} = 1$, guard for $t_{AS}$ is false, so return is blocked. With $\mathsf{timeout}_A$ eventually true, $t_{AR}$ becomes enabled and by Lemma 5 must fire within its deadline. ■

## *E. Governance Lemmas*

### *1) Lemma 10 (Unsafe conditions enable governance)*

**Statement.** If $\mathsf{UR}(\sigma) = 1$, then from any mode state $X \in S, M, A$, there exists a strongly-timed enabled path to $P_R$with bounded total time (possibly a single transition $t_{XR}$).

**Proof.** By SMARt construction, $\mathsf{UR}$is a guard enabling governance transitions $t_{SR}$, $t_{MR}$, and $t_{AR}$(or bounded composition $S \to M \to R$). Apply Lemma 5 along the enabled path. ■

### *2) Lemma 11 (Absorbing governance without external authorization)*

**Statement.** If all transitions out of $P_R$require an external authorization predicate $\mathsf{ext}_{\mathsf{a}}\mathsf{uth} = 1$and safety predicates, then while $\mathsf{ext}_{\mathsf{a}}\mathsf{uth} = 0$(or while unsafe predicates hold), no transition leaving $P_R$is enabled.

**Proof.** Guards on all outgoing transitions are false, so they are disabled. ■

## *F. Corollaries: Propositions 1-5*

### *1) Corollary 1 (Proposition 1: Bounded Autonomy)*

**Follows from:** Lemmas 2, 7. If $\mathsf{invalid}$persists in $P_S$, token leaves $P_S$within $\Delta_S$. ■

### *2) Corollary 2 (Proposition 2: Formal Bound on Ungrounded Generations)*

**Follows from:** Lemmas 3-4 plus Corollary 1 (optional).
Output requires token in $P_S$, but invalidity forces exit from $P_S$and/or output guards disable output under invalidity. Hence output under invalidity is structurally precluded. ■

### *3) Corollary 3 (Proposition 3: Mandatory Escalation)*

**Follows** Lemma 8. Token cannot remain indefinitely in $P_M$; it returns to $P_S$ or escalates to $P_A/P_R$ under bounded budgets. ■

### *4) Corollary 4 (Proposition 4: Governance Reachability)*

**Follows** Lemmas 10-11. Unsafe conditions reach $P_R$in bounded time; $P_R$is absorbing without authorization. ■

### *5) Corollary 5 (Proposition 5: Distributed Soundness)*

**Follows** Lemma 9. Persistent disagreement blocks return to $P_S$and triggers escalation to $P_R$. ■